# Diagnosis of Alzheimer's Disease via Multi-modality 3D Convolutional Neural Network

Yechong Huang[1], Jiahang Xu[1], Yuncheng Zhou[1], Tong Tong[2], Xiahai Zhuang[1]*, and the Alzheimer's Disease Neuroimaging Initiative

[1] School of Data Science, Fudan University, Shanghai, China

[2] Fujian Provincial Key Laboratory of Medical Instrument and Pharmaceutical Technology, Fuzhou, China

* Corresponding author: Xiahai Zhuang, zxh@fudan.edu.cn

## Abstract

Alzheimer's Disease (AD) is one of the most concerned neurodegenerative diseases. In the last decade, studies on AD diagnosis attached great significance to artificial intelligence (AI)-based diagnostic algorithms. Among the diverse modality imaging data, T1-weighted MRI and $^{18}$F-FDG-PET are widely researched for this task. In this paper, we propose a novel convolutional neural network (CNN) to fuse the multi-modality information including T1-MRI and FDG-PDT images around the hippocampal area for the diagnosis of AD. Different from the traditional machine learning algorithms, this method does not require manually extracted features, and utilizes the state-of-art 3D image-processing CNNs to learn features for the diagnosis and prognosis of AD. To validate the performance of the proposed network, we trained the classifier with paired T1-MRI and FDG-PET images using the ADNI datasets, including 731 Normal (NL) subjects, 647 AD subjects, 441 stable MCI (sMCI) subjects and 326 progressive MCI (pMCI) subjects. We obtained the maximal accuracies of 90.10% for NL/AD task, 87.46% for NL/pMCI task, and 76.90% for sMCI/pMCI task. The proposed framework yields comparative results against state-of-the-art approaches. Moreover, the experimental results have demonstrated that (1) segmentation is not a prerequisite by using CNN, (2) the hippocampal area provides enough information to give a reference to AD diagnosis.

## 1. Introduction

The aging of global population results in increasing number of people with dementia. Recent studies indicate 50 million people are living with dementia [1], of whom 60%-70% have Alzheimer's Disease (AD) [2]. Known as one of the most common neurodegenerative diseases, AD can result in severe cognitive impairment and behavior issue.

With the progression of AD, the structure and metabolic rate of the brain change. The phenotype includes the shrinkage of cerebral cortices and hippocampi, the enlargement of ventricles, and the change of regional glucose uptake. These changes can be quantified with the help of medical imaging techniques such as magnetic resonance imaging (MRI), positron-emission tomography (PET) and computed tomography (CT) [3]. For instance, T1-weighted MRI provides high-resolution structural information of the brain, enabling the measurement of structural metrics such as thickness, volume and shape. Meanwhile, $^{18}$F-FDG-PET indicates the regional cerebral metabolic rate of glucose, making it possible to evaluate the metabolic activity of tissues. By computing and analyzing from these medical images, one can obtain important reference to assist the diagnosis and the prediction of AD [4].

Mild cognitive impairment (MCI) is a neurological disorder which involves cognitive impairments with minimal impairment in instrumental activities of daily living [5]. MCI involves the onset and evolution of cognitive impairments beyond a normal expectation of an individual's age and education, but which are not significant enough to interfere with her or his daily activities [6]. It may occur as a transitional stage between normal aging and the preclinical phase of dementia. MCI is a large heterogeneous group because of its variety of clinical outcomes [7]. In this work, we classified MCI into progressive mild cognitive impairment (pMCI) and stable mild cognitive impairment (sMCI), which are retrospective diagnostic terms based on the clinical follow-up. As pMCI refers to the MCI subjects who converted to dementia according to the DSM-IV criteria during the follow-up [8], sMCI is defined when the subjects do not fulfill the criteria. Identification between pMCI and sMCI plays an important role for the early diagnosis of dementia, which can assist clinicians to propose effective therapeutical interventions of the disease process [9].

This work aims at differentiating AD or potential AD patients from normal subjects accurately and automatically using medical images around the hippocampal area and recent technologies in deep learning. This facilitates a fast-preclinical diagnosis. The method is further extended for the classification between sMCI and pMCI so that an early diagnosis of dementia would be possible.

In our proposed method, multi-modality data were used. We used two modalities in our experiments including the T1-weighted MRI and $^{18}$F-FDG-PET. Up to now, numerous studies have been published on diagnosing AD by utilizing T1 MRI Sorensen et al. segmented the brains and extracted features like the thicknesses of cortices and volumes of tissues in the selected regions of interest (ROIs) [10]. After that they used linear discriminant analysis (LDA) to classify AD, MCI and normal (NL), and achieved 63.0% classification accuracy on the CADDementia dataset. David et al., implemented kernel metric learning in classification [11]. Based on the segmentation results provided by FreeSurfer, morphological features like cortical volume, surface area and thickness were extracted and utilized for further classification. In the end they obtained an overall accuracy of 60.1% on the CADDementia dataset. Another popular machine learning method is random forest. Lebedeva et al. extracted the cortical thickness and subcortical volumes as structure features of MRI and used mini-mental state examination (MMSE) as a cognitive measure [12]. Ardekani et al. took hippocampal volumetric integrity of MRI and neuropsychological scores as the selected features [13]. Both studies used random forest to implement the classification because random forest has been demonstrated as a feasible and reliable machine learning method for classification.

The imaging data from $^{18}$F-FDG-PET are widely used to observe metabolic processes in the body, including for diagnosis of AD. Silveira et al., proposed a boosting learning method which used a mixture of simple classifiers to perform voxel-wise feature selections. They achieved an accuracy of 90.97% in the detection of AD and 79.63% of MCI on the ADNI dataset [15]. Cabral et al., used favorite class ensembles to form ensembled SVM and random forest, and achieved 66.78% accuracy in AD, NL and MCI 3-class classification problem on the ADNI dataset [16].

In addition to the single modality classifications, taking both T1-weighted MRI and $^{18}$F-FDG-PET together is also a main concern of AD diagnosis. Gary et al., took regional MRI volumes, voxel-based PET intensities, cerebrospinal fluid (CSF) biomarkers and genetic information as features and used random-forest-based similarity measures to classify AD and MCI from NL, and got 89% accuracy for AD/NL, and 75% for MCI/NL on the ADNI dataset [17]. Besides, Zhang et al., conducted classification based on MRI, PET and CSF biomarkers [18]. Using SVM, they obtained 93.2% accuracy for AD/NL, and 76.4% for MCI/NL. Moreover, other imaging modalities or PET tracers can be taken into account, like Rondina et al., used T1-weighted MRI, $^{18}$F-FDG-PET and rCBF-SPECT as imaging modalities while Wang et al., took $^{18}$F-FDG and $^{18}$F-florbetapir as tracer of PET [19,20].

The studies mentioned above follow four steps in the diagnosis algorithms, namely data preprocessing, segmentation, feature extraction and classification. During data preprocessing, methods of image correction, normalization and coordinates transformation are applied to the acquired data so that the image quality and consistency are ensured. During segmentation, data are manually or automatically partitioned into multiple segments based on anatomy or physiology. In this way, the regions of interest (ROIs) are well-defined, making it possible to extract features from ROIs. Finally, these features will be fed into the classification step so that classifiers are able to learn useful diagnostic information and propose predictions of AD for given test subjects. Though highly reliable and explainable, these steps could be sophistic and lowly integrated, as different platforms are used in different steps of these algorithms. In addition, the automatic and accurate segmentation of brain is still an open question, leading to more uncertainty of the extracted features.

Benefited from the rapid development of computer science and the accumulation of patients' data, deep learning has become a popular and useful technique in the field of medical imaging recently. Therefore, we propose to use deep learning framework in our work to implement the feature extraction and classification steps. The general applications of deep learning in medical imaging are mainly about feature extraction, image classification, object detection, segmentation and registration [21]. Among the deep learning networks, convolutional neural networks (CNNs), recurrent neural networks (RNNs), deep belief networks (DBNs), auto-encoders (AEs) and restricted Boltzmann machines (RBMs) are common choices. Take CNN-based AD diagnosis as an instance, different people used different architectures. Hosseini-Asl et al. built a 3D-CNN based on a 3D convolutional auto-encoder, which takes fMRI images as input and gives prediction of AD/MCI/NL [22], while Sarraf et al. used a LeNet-5-like CNN to classify AD from NL based on fMRI [23]. Multi-modality classification was also implemented in CNN. Liu et al. conduced a T1-MRI+FDG-PET based cascaded CNN, which utilized a 3D CNN to extract features respectively for each modality and adopted another 2D CNN to combine multi-modality features for task-specific classification [24].

In this work, we propose a multi-modality AD classifier, which takes MRI and PET images around the hippocampal area as input, and give predictions in the NL/AD task, the NL/pMCI task and the sMCI/pMCI task. The main contributions of our work are listed below:

(1) We showed that segmentation of the key substructures, such as hippocampi, is not a necessary step in CNN-based classification.
(2) We showed that a ROI enclosing the hippocampal area provides enough information to give reference to AD diagnosis.
(3) We conducted a 3D VGG variant CNN to implement single modality AD diagnosis.
(4) We introduced a new framework to fuse complementary information from multiple modalities in our proposed network, for the classification tasks of NL/AD, NL/pMCI and sMCI/pMCI.

The rest of the paper is organized as follows. Section 2 describes the datasets. Section 3 presents the classification algorithm of CNN and VGG neural network. In Section 4, we elaborate on the methodologies of the VGG variant networks for the different classification tasks and provide the results. Section 5 presents the discussion and conclusion of this work.

## 2. Material

In this section, we presented the multi-modality image datasets that were used in this study, the data pre-processing methods. MRI and PET are two essential modalities that are widely used for AD diagnosis. Biomarkers of Alzheimer's disease (AD) extracted from MRIs is an active research area, whose importance in AD was underlined by its inclusion in criteria for AD diagnosis. In particular, the decrease of hippocampal volume is one of the best-established biomarkers used in research studies to stage the progression of Alzheimer's disease (AD) pathology in the brain of patients across the spectrum of the disease [25]. Therefore, the hippocampi are the most studied region of AD, which are so far the only MRI biomarker that has been qualified for enrichment of clinical trials [26]. For PET images, published studies indicate that AD may cause the decline of 18-Fluoro-DeoxyGlucose (FDG) uptake in the temporoparietal cortex, as well as brain atrophy in the medial, basal, lateral temporal lobes, the medial and lateral parietal cortices. As a result, the ROI of metabolic changes in cortices, especially temporal lobes with FDG biomarkers may help AD diagnosis [27,28,29]. When dealing with PET images, we tried different ROIs containing hippocampi and cortices.

### 2.1 Image Acquisition

The datasets we used in this work were downloaded from the Alzheimer's Disease Neuroimaging Initiative (ADNI) database, which is publicly available on the website http://adni.loni.usc.edu. ADNI is a longitudinal multicenter study designed for the AD detection at the earliest possible stage (pre-dementia) and tracking of progression of Alzheimer's disease (AD) with biomarkers. The study comprises four main phases, the initial five-year study (ADNI-1) was extended by two years in 2009 by a Grand Opportunities grant (ADNI-GO), and in 2011 and 2016 by further competitive renewals of the ADNI-1 grant (ADNI-2, and ADNI-3, respectively). The participant subjects were recruited across North America during each phase of the study, agreed to complete a variety of imaging and clinical assessments, and the results are shared by ADNI through the USC Laboratory of Neuro Imaging's Image and Data Archive (IDA). ADNI helps researchers to collect, validate and utilize data, including magnetic resonance imaging (MRI) and Positron Emission Tomography (PET) images, genetics, cognitive tests, CSF and blood biomarkers as predictors of the disease, to measure and track the progression of mild cognitive impairment (MCI) and early Alzheimer's disease (AD).

**Table 1** Summary of the studied subjects of MALPEM dataset.

| Diagnosis | Number | Age | Gender(M/F) | MMSE |
|---|---|---|---|---|
| AD | 1355 | 76.13±7.50 | 772/583 | 21.89±4.33 |
| NL | 1506 | 76.04±5.81 | 776/730 | 29.04±1.20 |

**Table 2** Summary of the studied subjects from the Paired dataset.

| Diagnosis | Number | Age | Gender(M/F) | MMSE |
|---|---|---|---|---|
| AD | 647 | 76.36±7.21 | 361/287 | 24.84±2.65 |
| pMCI | 326 | 75.00±7.06 | 212/114 | 27.22±1.74 |
| sMCI | 441 | 74.37±7.40 | 297/144 | 28.15±1.55 |
| NL | 731 | 76.16±6.02 | 421/310 | 28.99±1.20 |

In this work, we used the T1-weighted magnetic resonance (MR) imaging data and the 18-Fluoro-DeoxyGlucose PET (FDG-PET) imaging data from the baseline and follow-up visit in ADNI. The MRI images were acquired from volumetric 3D Accelerated MPRAGE and SPGR, and the FDG-PET images were acquired 30–60 min post-injection. To integrate the PET images for each acquisition, we calculated the averaged intensity of images after intensity normalization with different post-injection time to generate a single PET image in this acquisition. One can visit the ADNI website for more details about the acquisition steps of MRI and PET [30].

We used two datasets in this work. To verify the effect of segmentation on classification, a larger dataset containing only MRI images, called the MALPEM dataset [31], and a smaller dataset containing MRI and corresponding PET images, called Paired dataset were chosen.

In the MALPEM dataset, T1-weighted images were segmented into 138 anatomical regions using multi-atlas label propagation with expectation-maximization (MALPEM), which is a software package to perform whole-brain segmentation of T1-weighted MRI images, available on the website https://biomedia.doc.ic.ac.uk/software/malp-em/ [31]. In all, 2861 subjects containing both MRI images and segmentation labels were obtained, including AD and NL. All subjects we studied are summarized in **Table 1**.

About the Paired dataset, we took the following steps to screen out the corresponding MRI and PET images of the patient taken in the same period. Firstly, for each patient, we picked up all the corresponding MRI images and FDG-PET images. For each image, the data column contains the patient ID and the acquisition date, by which we can determine the time interval between different MRI and PET images. Then, we merged MRI and PET images for the same patient and combined the nearest MRI and PET images according to the acquisition date. MRI and PET images were regarded as paired multi-modality images if the two images were taken within one year. After that, we manually checked the patients' diagnosis to ensure the two images point to same diagnosis. After removing pairs of MRI and PET images captured with interval more than one year, we got the remaining MRI and PET images as subjects with multi-modality images as the Paired dataset for classification. After data filtering described above, 647 AD subjects, 1391 MCI subjects and 731 NL subjects were acquired.

The MCI subjects were classified into progressive mild cognitive impairment (pMCI) and stable mild cognitive impairment (sMCI) in this work. According to the judging criteria we mentioned before, MCI is defined as pMCI if it is developed to AD within three years for the current MCI patient, or defined as sMCI if it is not developed to AD for at least three years. Besides, subjects without follow-up data for more than three years were ignored. Through this standard, we identified 326 pMCI subjects and 441 sMCI subjects.

Finally, we acquired imaging data from 1211 ADNI participants including 647 AD, 767 MCI (326 MCI converters (pMCI) and 441 MCI non-converters (sMCI)), 731 NL subjects. All the subjects we studied are summarized in **Table 2**.

## 2.2 Data Processing

The pre-processing of MRI and PET images was implemented by zxhtools, which is available on the website http://www.sdspeople.fudan.edu.cn/zhuangxiahai/0/zxhproj/. Based on this platform, several image registration and segmentation tools have been developed. In this work, zxhreg and zxhtransform were mainly used for the registration of the radiological images. Before proceeding any further, all MRI images were re-oriented and resampled to the resolution of $221 \times 257 \times 221$ and the spacing of $1 \times 1 \times 1$ mm$^3$.

The significant position hippocampi hold could be stated from the description in the previous sections. Therefore, the hippocampal area in MRI and PET images were located to be the region of interest (ROI) in this work. In addition, due to the limitation of GPU RAM and computation ability in VGG network, we cropped the region of interest to represent the whole image, which is centered in the center point of hippocampi and dilate from it.

For the MALPEM datasets with segmentation result, we could directly calculate the center point of the hippocampi's center point. For the Paired dataset, we obtained the central points of MRI images through the following steps. Firstly, the location of the hippocampi is marked on MRI images in the MALPEM dataset, thus the center of hippocampi can be calculated. Secondly, one MRI image from the Paired dataset was selected as template of hippocampi's center point. The images from the MALPEM dataset were affine-registered to the template image, and the center points were averaged to identify the center point of hippocampi in template image. Finally, we registered the template image to other MRI images in the Paired dataset using affine-registration and used the corresponding affine matrix to determine the center point for each MRI image. Furthermore, each PET image was rigid-registered to respective MRI image for the identification of hippocampi's center point in PET images. After registration, PET images were transformed to a uniform isotropic spacing of $1 \times 1 \times 1$ mm$^3$.

After confirming the center point of ROI, we dilated and cropped ROI as big as $96 \times 96 \times 48$ voxels of MRI images from the center point of hippocampi. In the MALPEM dataset, in order to verify the effect of segmentation on classification, we processed the cropped ROI and corresponding labels through three different treatments and obtained three datasets: Raw, WithSeg and Bin. The Raw dataset contains MRI raw images only, which received all the imaging information of the hippocampi and surrounding areas. The WithSeg dataset is made up of MRI images masked by binary labels, it takes original images and segmentation results as input. The Bin dataset is made up of binary hippocampi segmentation labels, only indicating information about the shape and volume of the hippocampi. Through the comparison of classification performance using these three datasets, it can show whether the segmentation results have an important effect on AD diagnosis. The three datasets are shown in **Figure 3 A, B, C**.

When it comes to PET images in the Paired dataset, we used two different ways to generate the patch of PET images. The dataset generated by the first way is called the Origin dataset, using exactly the same ROI with the MRI, which means that voxels of PET have the same spacing with corresponding origin MRI. The dataset generated by the second way is called the Dilated dataset, which has the same ROI center and orientation with corresponding MRI but has $2 \times 2 \times 2$ mm$^3$ spacing. In other words, the Dilated dataset has 8 times of the volume but a lower spatial resolution, and the entire temporal lobes of both sides are included.

After data processing, the dataset was randomly split into the training set and the testing set according to patient ID to ensure that all subjects of the same patient only appear in the training set or the testing set at the same time. Finally, 70% of the dataset were used as the training set, 10% as the validation set, and 20% as the testing set by random sampling.

## 3. Methodology

In this section, we introduce the proposed classification algorithm of a convolution-based neural network in detail. Convolutional neural network (CNN) [32] is a deep feedforward neural network composed of multi-layer artificial neurons, having excellent performance in large-scale image processing. Unlike traditional methods of manually extracted features of radiological images, CNNs are used to learn general features automatically. CNNs are trained with back propagation algorithm, usually consisting of multiple convolutional layers, pooling layers and fully connected layers, connecting to output unit through full connection layers or other methods. Compared to other deep feedforward networks, CNNs have fewer connections and smaller number of parameters due to the weight of the convolution kernel sharing and are therefore easier to train and more popular.

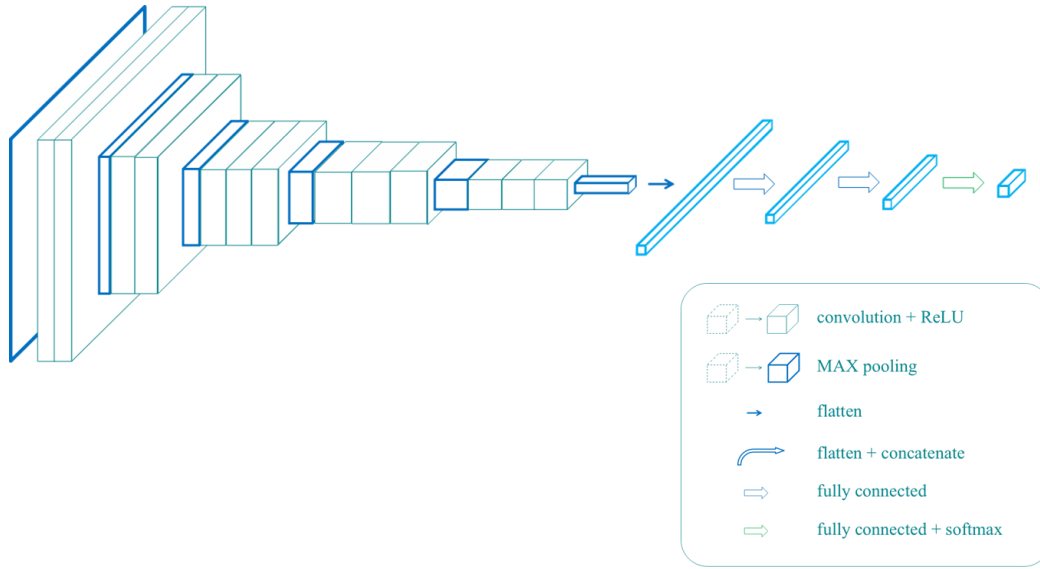

**Figure 1** VGG16, which is a very deep network consisting of 13 convolution layers, 5 max pooling layers and 3 fully connected layers. Several convolution layers are followed by max pooling layer, reducing the dimensionality. In this figure, the original image with pixel size 224 × 224 formed a feature map of size 7 × 7 × 512 after multiple convolutions and pooling layers, obtaining classification result after fully connected and Softmax layers.

With CNNs becoming popular in the computer vision field, a number of attempts have been made to improve the original network structure to achieve better accuracy. VGG [33] is a neural network based on improved AlexNet [34] in 2014, achieved a 7.3% Top-5 error rate in the 2014 ILSVRC competition [35]. VGGs further deepen the network based on AlexNet by adding more convolutional layers and pooling layers. A demonstration of VGG16 is shown in **Figure 1**, it is a very deep network consists of several convolution layers followed by max pooling layers. Different from traditional CNNs, VGGs evaluate very deep convolutional networks for large-scale image classification, which come up with significantly more accurate CNN architectures and can achieve excellent performance even when used as a part of a relatively simple pipelines.

## 4. Experiments

In this section, we used several parts of work as shown below to introduce our experiment. In Section 4.2, we used different types of data to determine the proper datatype and ROI by two experiments. In Section 4.2.1, the classification results by datasets with/without segmentation were compared to show whether segmentation is necessary or not in CNN. In Section 4.2.2, different spacings of PET images were tested, and we found that the classification model using a small spacing and a small region has similar performance with a large spacing and a large region. In Section 4.3, we constructed a VGG-like multi-modality AD classifier, which took both T1-MRI and FDG-PET data as input and gave a prediction. In Section 4.4, we trained and tested our network with pMCI and sMCI data. Finally, Section 4.5 compared our method with state-of-the-art methods.

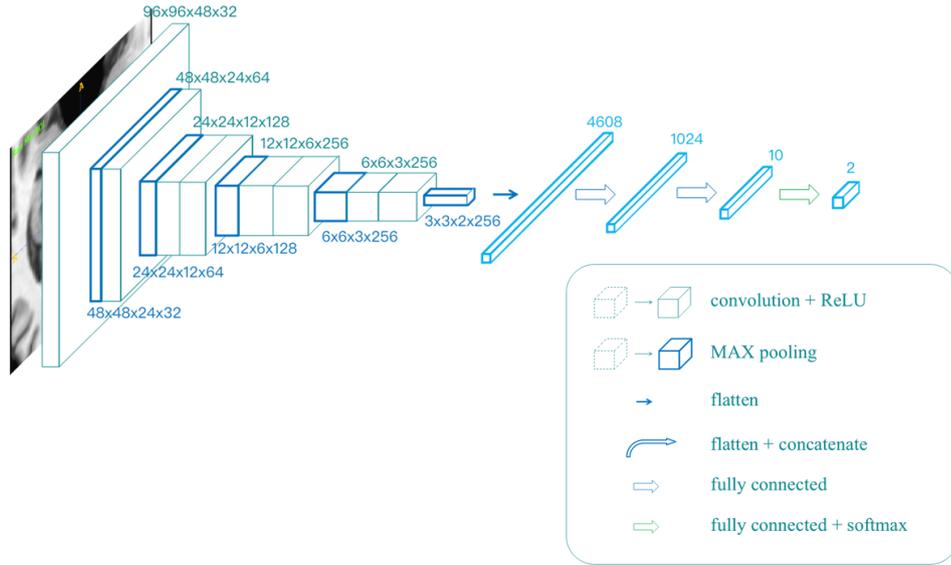

**Figure 2** The architecture of the single modality classifier.

### 4.1 Experimental setup

All the networks mentioned above were constructed with the help of TensorFlow [36] on Python platform. Training procedures of the networks were conducted on a PC with Nvidia GTX1080Ti GPU. During training, batch normalization [37] was deployed in convolution layers and dropout [38] was deployed in full-connection layers to avoid overfitting. To accelerate the training process and to avoid local minimums, we used ADAM algorithm [39] as the training optimizer. Batch size was set to 16 when training single modality networks and 8 when training multi-modality networks. In training it took one to several minutes per epoch, with respect to different datasets and network architectures, as an ordinary training schedule contains 150 epochs. In most cases, the training loss would converge in 30 epochs. During training, the parameters of the networks were saved every 10 epochs. These parameters were used to run on the validation set later. During the validation, accuracies and the receiver operating characteristic (ROC) curves of the classification were calculated, and the parameters with the highest accuracy and area under ROC (AUC) was chosen to be the final classifier.

### 4.2 Data Types Analysis

In order to determine the proper data type for network to train, we designed two experiments to evaluate the classification performance of models when feeding with different datatypes in this part, as listed below:
(1) Testing whether segmentation is needed in the MRI images. We used three different datasets, with or without segmentation, to show that segmentation is not necessary in CNN.
(2) Finding a proper PET ROI. Different spacings of PET images were tested, as we found that the classification model using a small spacing and a small region has similar performance of that with a large spacing and a large region.

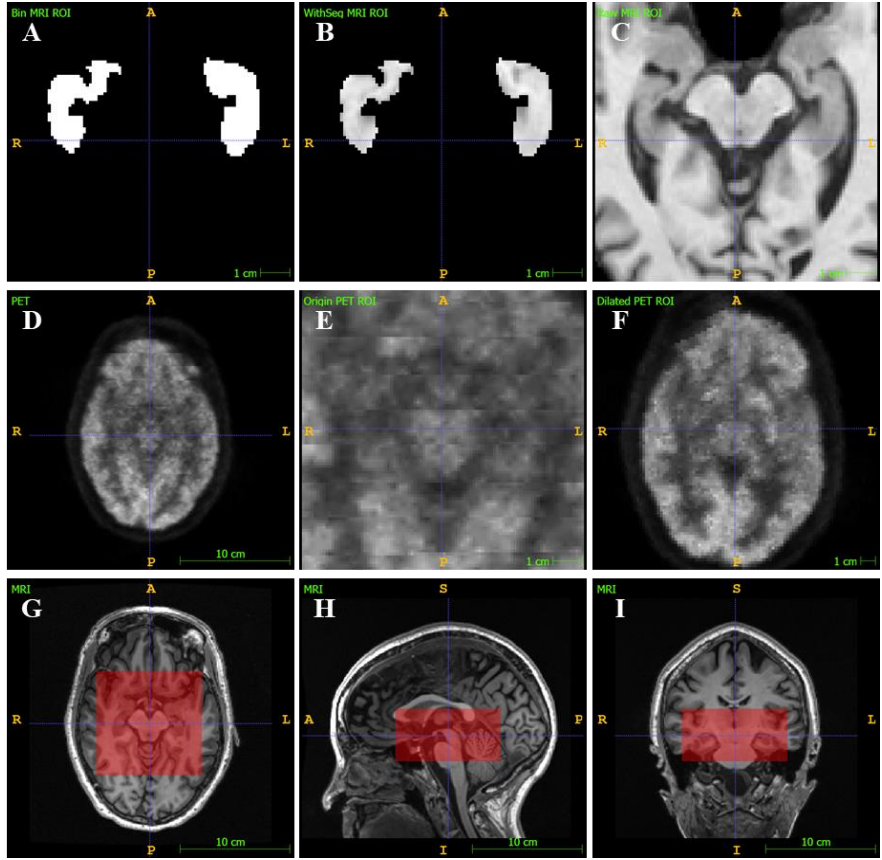

**Figure 3** Demonstrations of the datasets and ROI. A, B, C are generated from the same MRI image to demonstrate the Bin (A), WithSeg (B), and Raw (C) dataset. A is a binary image of the segmentation of hippocampi. B is a raw image masked by hippocampal segmentation. C is a cropped raw image. D, E, F are generated form the same PET image to demonstrate the Origin (E) and Dilated (F), as D is the PET image. E is cropped from D, and F is down sampled from D. G, H, I demonstrates the selected ROI, as G is horizontal plane, H is sagittal plane, and I is coronal plane.

All the models mentioned above were trained in the same network, as shown in **Figure 2**. The input resolution was $96 \times 96 \times 48$ voxels, and the network contained 8 convolution layers, 5 pooling layers and 3 full connection layers. The output was given through the softmax function.

### 4.2.1 Influence of Segmentation

In this part, we compared the classification results using datasets with/without segmentation to show that segmentation is not necessary in CNN. As mentioned above, segmentation takes an important role in traditional classification methods, for it labels voxels with anatomy meanings and defines ROIs. However, segmentation is also known for time-consuming. Meanwhile, it's possible that CNN can extract useful features directly from raw images, as lots of CNNs show a strong ability of locating keypoints in object detection tasks in natural image processing [40, 41]. However, segmentation can help in many ways, including locating ROIs, specifying the morphologic and volumetric features, blocking out confusing tissues, and speeding up the convergence when training the models.

**Table 3** Summary of the models trained from the Bin, WithSeg and Raw datasets for NL/AD task. The MALPEM dataset was used.

| MRI ROI | ACC | SEN | SPE | AUC |
|---|---|---|---|---|
| **Bin** | 76.57% | 83.87% | 71.51% | 84.24% |
| **WithSeg** | 79.21% | 76.61% | 81.01% | 84.63% |
| **Raw** | 84.82% | 87.90% | 82.68% | 87.47% |

**Table 4** Summary of the models trained from the Origin PET and the Dilated PET datasets for NL/AD task. The Paired dataset was used.

| PET ROI | ACC | SEN | SPE | AUC |
|---|---|---|---|---|
| **Origin PET** | 89.11% | 90.24% | 87.77% | 92.69% |
| **Dilated PET** | 89.44% | 87.20% | 92.09% | 90.35% |

We segmented the AD and NL subjects of T1-MRI images with MALP-EM algorithm [31] and obtained the MALPEM datasets, including 2861 subjects that contain both MRI images and segmentation. In our assumption, segmentation itself indicates the shapes, volumes, textures and relative locations of hippocampal areas. Therefore, the data obtained from the subjects was generated three different datasets, as shown in **Figure 3 A, B, C**. The Raw dataset contains MRI raw images only; the Bin dataset is made up of binary hippocampal segmentation labels and the WithSeg dataset is made up of MRI images masked by binary labels.

For each model trained from these datasets, accuracy and AUC were calculated, as listed in **Table 3**. Among all three models, the model trained from the Raw dataset performed best, followed by the WithSeg dataset. Model trained from the Bin dataset also gave favorable prediction, though inferior to the Raw dataset and the WithSeg dataset. The results indicate that segmentation do contain information needed for classification. However, segmentation is not necessary for the classification task since CNN is able to learn useful features without labeling the voxels. In addition, features from region out of hippocampi also add further information in separating AD patients apart from normal ones.

### 4.2.2 PET ROI Determination

In this part, we discussed the ROI selection of PET images. We tested different spacings of PET images to show which ROI is better for PET in classification. Due to the limitation of GPU RAM and computation ability, it was difficult to take the entire image as the network input, as our proposed network only took a region as big as $96 \times 96 \times 48$ voxels, which was still 2.91 times of the input size of original VGG ($224 \times 224$ pixels $\times$ 3 channels). Hence, the selection of ROI took an important role, as only the features in the ROI would be taken. When regarding to MRI images, the selection of ROI was with little doubt, for the hippocampal area was long to be the main concern of AD research [26, 27]. However, the ROI of PET images varied, as studies also attached significance to metabolic changes in cortices, especially temporal lobes [28, 29].

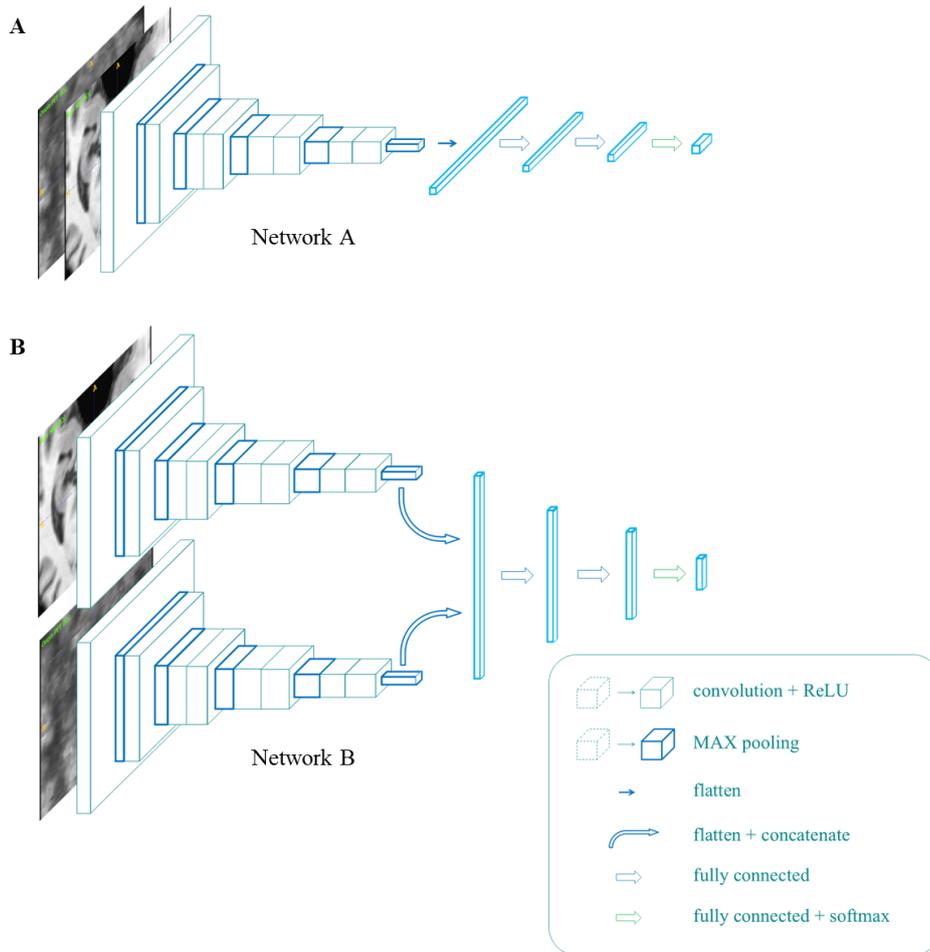

**Figure 4** The architecture of the multi-modality network A and B.

To verify the effect of ROI containing temporal lobes on classification, we generated two datasets from the PET images, the Origin dataset and the Dilated dataset, as shown in **Figure 3 D, E, F**. The Origin dataset uses exactly the same ROI with the MRI ROI, which means that voxels of a pair of MRI and PET images have the same spacing and the same orientation and are voxelwisely aligned. The Dilated dataset has the same ROI center and orientation with the MRI ROI but has a $2 \times 2 \times 2$ mm$^3$ spacing, in which the entire temporal lobes of both sides are included. In short, the Dilated dataset has 8 times of the volume but a lower spatial resolution than the Origin dataset.

Two models were trained from these two datasets, and metrics were calculated, as listed in **Table 4**. As is shown, the performance of two models were relatively close. Although the Origin dataset has higher spatial resolution, the Dilated dataset contains more features. Further, when considering multi-modality classification tasks, the Origin dataset might be better, because PET images in the Origin dataset were voxelwisely aligned with paired MRI images, which could help better locate spatial features. Therefore, we chose the same ROI in MRI and PET images, as the regions shown in **Figure 3 G, H, I**.

**Table 5** Summary of the models trained from single modality protocols and three multi-modality protocols for NL/AD task. The Paired dataset was used.

| Method | ACC | SEN | SPE | AUC |
|---|---|---|---|---|
| **MRI** | 81.19% | 79.27% | 83.45% | 83.67% |
| **PET** | 89.11% | 90.24% | 87.77% | **92.69%** |
| **A** | 87.79% | 85.98% | **89.93%** | 89.42% |
| **B1** | **90.10%** | **90.85%** | 89.21% | 90.84% |
| **B2** | 89.44% | 89.02% | **89.93%** | 92.01% |

**Table 6** Summary of the models trained from three multi-modality protocols for NL/AD task, NL/pMCI task and sMCI/pMCI task. The Paired dataset was used.

| Method | A | | | | B1 | | | | B2 | | | |
|---|---|---|---|---|---|---|---|---|---|---|---|---|
| | ACC | SEN | SPE | AUC | ACC | SEN | SPE | AUC | ACC | SEN | SPE | AUC |
| **NL/AD** | 87.79% | 85.98% | 89.93% | 89.42% | **90.10%** | **90.85%** | **89.21%** | **90.84%** | 89.44% | 89.02% | 89.93% | 92.01% |
| **NL/pMCI** | 70.49% | 73.17% | 65.00% | 71.63% | 79.10% | 87.80% | 61.25% | 76.84% | **82.38%** | **87.20%** | **72.50%** | **81.64%** |
| **sMCI/pMCI** | 65.28% | 65.63% | 65.00% | 65.81% | 65.28% | 54.69% | 73.75% | 66.82% | **72.22%** | **73.44%** | **71.25%** | **77.49%** |

### 4.3 Multi-modality AD Classifier

In this part, we constructed a VGG-like multi-modality AD classifier and compared the results with single modality models. When using single modality, the information that a classifier can obtain is limited, as one medical imaging method can only profile one or several aspects of AD pathological changes, which is far to be complete. For example, T1-MRI images provide high-resolution brain structure but give little information about the functional information about the brain. Meanwhile, FDG-PET images are fuzzy, but do better in revealing the metabolic activity of glucose in the brain. In order to take as more information of the brain as possible, we introduced a classification framework to fuse multi-modality information.

To prepare the dataset, we firstly matched MRI with PET images and translated them into same world coordinates. After that, paired images of MRI and PET were aligned by rigid registration to ensure that voxels of the same indices in the paired images represent the same part of the brain. After that paired images were cropped according to the center point of MRI images, the Paired dataset were obtained.

To implement the multi-modality classifier, we proposed two different network architectures, as shown in **Figure 4**. In **Figure 4 A**, MRI and PET images were used as two parallel channels, in which way paired images were stacked into 4D images. In these 4D images, the first three dimensions represent the three spatial dimensions, and the fourth dimension represents different channels. In **Figure 4 B**, MRI and PET images have separate entrances, as they are convolved respectively in two separate VGG-11s, and the extracted features are concatenated together. This network was trained in two protocols, denoted by B1 and B2. In B1, model was trained with the weights shared for the convolutional layers. Meanwhile in B2, the weights of two VGG-11s were updated separately in training.

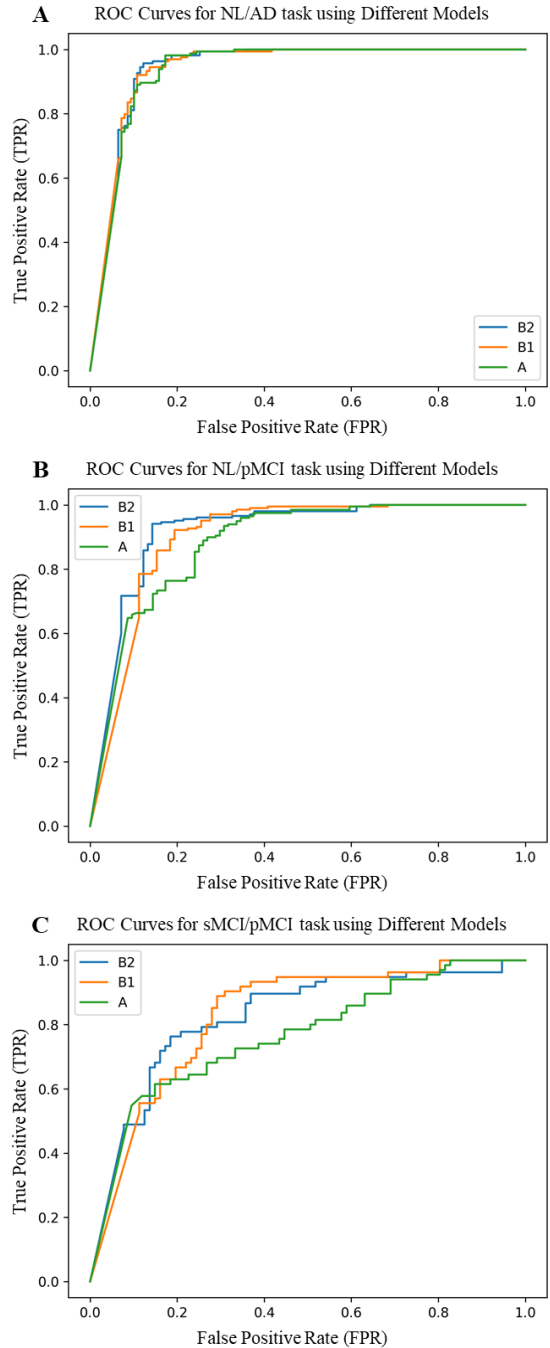

**Figure 5** ROC curves of different models. A, B, C are ROC curves for three tasks using different models. A shows the ROC curve for NL/AD task using model trained from protocol A, B1 and B2, while B shows the ROC curve for NL/pMCI task, C shows the ROC curve for sMCI/pMCI task, respectively.

We trained five models based on paired MRI and PET images, as two single modality models, MRI and PET, and three multi-modality models, A, B1 and B2. The results are shown in **Table 5** and **Figure 5 A**. As shown in **Table 5**, multi-modality classifiers have better performance than single modality classifiers. Besides, among three multi-modality models, model trained with protocol B1 has the highest accuracy and sensitivity, while model trained with protocol B2 has the highest specificity and AUC.

Table 7 Comparison of our proposed method and Liu's multi-modality method.

| Method | Subjects | Modality | NL vs. AD | | | | NL vs. pMCI | | | |
|---|---|---|---|---|---|---|---|---|---|---|
| | | | ACC | SEN | SPE | AUC | ACC | SEN | SPE | AUC |
| Liu et al. 2018 | 93 AD + 204 MCI + 100 NL | MRI | 84.97% | 82.65% | 87.37% | 90.63% | 77.84% | 76.81% | 78.59% | 82.72% |
| | | PET | 88.08% | 90.70% | 85.98% | 94.51% | 78.41% | 77.94% | 78.70% | 85.96% |
| | | Both | **93.26%** | **92.55%** | **93.94%** | **95.68%** | 82.95% | 81.08% | **84.31%** | 88.43% |
| Proposed method | 465 AD + 567 MCI + 480 NL | MRI | 81.19% | 79.27% | 83.45% | 83.67% | - | - | - | - |
| | | PET | 89.11% | 90.24% | 87.77% | 92.69% | - | - | - | - |
| | | Both | 89.02% | 89.93% | 92.01% | 89.02% | **87.46%** | **90.73%** | 80.61% | **90.31%** |

## 4.4 Classification of sMCI/pMCI and NL/pMCI tasks

In this part, we trained and tested our network with pMCI and sMCI subjects. Simply classifying AD patients from normal controls is relatively easy, but doesn't have much guidance significance, because the development of AD can be spotted by the behavior changes of patients. In addition, there are lots of alternative indicators in diagnosis. Therefore, the preclinical diagnosis of AD seems to be more meaningful, as one of the main concerns is telling progressive MCI from stable MCI and normal controls. As pMCI would progress to AD, classifying pMCI could give a prediction of the development of MCI, and thus have high reference value and clinical meaning.

According to Lin et al. [42], the models that trained from the NL/AD training set perform better than the models trained from the sMCI/pMCI training set in the sMCI/pMCI task. Therefore, we trained models with the NL/AD training set and tested the models with the NL/pMCI testing set and the sMCI/pMCI testing set, as the results shown in **Table 6** and **Figure 5**. Though B1 performed slightly better in NL/AD task, B2 was superior in NL/pMCI and sMCI/pMCI tasks. These results indicate that features of MRI and PET turns to be more consistent when dementia is highly developed, as convolutional kernels of model B1 shared weight while those of B2 didn't.

## 4.5 Comparison with Other Methods

In this part, we compared our method with those that were used in previous literature. We first compared our method with a state-of-the-art research using 3D CNN-based multi-modality models as well [42]. Liu et al. proposed a multi-modality cascaded CNN. They used the patch-based information of a whole brain to train or test their models and they integrated the information from the two modalities by concatenating the feature maps [43]. In this work, we used the hippocampal area to verify the connection between AD and the multi-modality structure of the hippocampi. **Table 7** shows the results of the two methods. Our method failed to outperform the cascaded CNN model in NL vs. AD task, but it achieved higher accuracy and AUC in the task of distinguishing pMCI from normal subjects. It should be noted that our models use the data from multiple facilities, our models take only the hippocampal area as an input. These would influence the behavior of our method.

Moreover, we compared our results with an extended CNN network using 2.5D (2D slices in different directions) MRI as an input. Lin et al., chose to reduce the amount of input by slicing the data instead of taking the hippocampi out as we did [42]. What's more, other multi-modality attempts were taken into account. Tong et al. used nonlinear graph fusion to join the features of different modalities [44]. In Zu et al.'s study, the feature selection from multiple modalities were treated as different learning tasks and a multi-kernel support vector machine (SVM) was adopted to fuse the selected features [45]. Liu et al. used stacked autoencoders (SAE) with a masking training strategy [43]. Jie et al. used manifold regularized multitask feature learning method to preserve both the relations among modalities of data and the distribution in each modality [46]. Li et al. used deep learning framework to predict the missing data [47]. **Table 8** compares the previous multi-modality models with our proposed models. Among all the results listed below, our results are favorable in the NL/AD task and are the best in the sMCI/pMCI task.

Table 8 Comparison of our proposed method and published AD diagnosis methods.

| Method | Subjects | NL vs. AD | | | | sMCI vs. pMCI | | | |
|---|---|---|---|---|---|---|---|---|---|
| | | ACC(%) | SEN | SPE | AUC | ACC(%) | SEN | SPE | AUC |
| Lin et al. 2018 | 93 AD + 204 MCI + 100 NL | 88.79% | - | - | - | 73.04% | - | - | - |
| Tong et al. 2017 | 37 AD + 75 MCI + 35 NL | 88.6% | - | - | 94.8% | - | - | - | - |
| Zu et al. 2016 | 51 AD + 99MCI + 52 NL | **95.95%** | - | - | - | 69.78% | - | - | - |
| Liu et al. 2015 | 85 AD + 168 MCI + 77 NL | 91.40% | 92.32% | 90.42% | - | - | - | - | - |
| Jie et al. 2015 | 51 AD + 99 MCI + 52 NL | 95.03% | - | - | - | 68.94% | - | - | - |
| Li et al. 2014 | 93 AD + 204 MCI + 101 NL | 92.87% | - | - | 89.82% | 72.44% | - | - | 70.14% |
| Proposed method | 465 AD + 567 MCI + 480 NL | 90.10% | 90.85% | 89.21% | 90.84% | **76.90%** | 68.15% | 83.93% | 79.61% |

## 5. Discussion and Conclusion

In this work, we proposed a VGG-like framework with several instances to implement a T1-MRI and FDG-PET based multi-modality AD diagnosing system. The ROI of MRI was selected to be the hippocampal area, as it is most frequently studied and is thought to be of the highest clinical value. Through the experiments, we proved that segmentation is not necessary in CNN, which is different from traditional machine learning based methods. However, registration is still needed, for the images we used were taken from different facilities and had different spacings and orientations. Although models obtained from the datasets Origin and Dilated had similar performance, the ROI of PET was chosen to be the same as MRI's, because the ROI of Origin was voxelwisely aligned with the ROI of paired MRI. In short, only hippocampal areas were used as ROIs in our proposed methods, which is the main difference between our study and the previous ones. Thus, we may construct a deeper network as well as feed it with higher resolution medical images, as we suppose that the hippocampal area itself can serve as a favorable reference in AD diagnosis.

**Table 9** Comparison of the performance of models trained from the NL/AD training set and the tasks' own training set. The Paired dataset was used.

| Task | Training Set | Testing Set | B1 | | | | B2 | | | |
|---|---|---|---|---|---|---|---|---|---|---|
| | | | ACC | SEN | SPE | AUC | ACC | SEN | SPE | AUC |
| **NL/pMCI** | NL/AD | NL/pMCI | 87.46% | 90.73% | 80.61% | 87.61% | 87.13% | 87.80% | 85.71% | 90.31% |
| **NL/pMCI** | NL/pMCI | NL/pMCI | 79.10% | 87.80% | 61.25% | 76.84% | 82.38% | 87.20% | 72.50% | 81.64% |
| **sMCI/pMCI** | NL/AD | sMCI/pMCI | 73.60% | 66.67% | 79.17% | 75.59% | 76.90% | 68.15% | 83.93% | 79.61% |
| **sMCI/pMCI** | sMCI/pMCI | sMCI/pMCI | 65.28% | 54.69% | 73.75% | 66.82% | 72.22% | 73.44% | 71.25% | 77.49% |

Since ROI was selected, we introduced multi-modality method to the classifier. Two networks and three types of models were proposed as listed in **Table 9**. Among these three types of models, the model trained from protocol B1, which means that MRI and PET images have separate inputs for convolutional layers, but shared convolutional kernels, performed the best in NL/AD task. One possible explanation is that MRI and PET images have some common features, and sharing weight helped models to extract these features during the training process. Furthermore, we used proposed networks to train NL/pMCI and sMCI/pMCI classifiers, both of them indicated the potential of preclinical diagnosis using our proposed methods.

We also followed Lin et al.'s lead and used the model trained by NL/AD subjects to distinguish sMCI and pMCI [42]. The results were better than that of the model trained by sMCI and pMCI themselves, as shown in **Table 9**. This is reasonable because the features of sMCI and pMCI are close to each other in the feature space and hard to differentiate while those of NL and AD are widely spread which makes the classification a lot easier. The same conclusion can be obtained by testing the NL/AD model on the NL/pMCI dataset. Specifically, when NL/AD model was used, the accuracy reached 76.90% for sMCI/pMCI and 87.46% for NL/pMCI, which were about 5 percent higher than the accuracy obtained using their own models. These results are also better than those of Lin et al.'s.

Regarding to the future work, first we only used two modalities, i.e., T1-MRI and FDG-PET, as input in this work. However, including new modalities can be easily implemented based on the proposed networks. The interested new imaging modalities incllude T2-MRI [48], $^{11}$C-PIB-PET [49], and other PET agents such as for amyloid protein imaging [50]. Also, the features extracted by CNN are hard for human beings to comprehend, while some methods like attention mechanism [51] are able to visualize and analyze the activation maps of the model, in which ways future work could be done to improve the classification performance and discover new medical imaging biomarkers.

To conclude, we have proposed a multi-modality CNN-based classifier for AD diagnosis and prognosis. VGG backbone, which is deeper than most similar studies, has been used and explored. The accuracy of models reached 90.10% for NL/AD task, 87.46% for NL/pMCI task and 76.90% for sMCI/pMCI task. As we only used hippocampal area as ROI, our work indicates that small but high-resolution ROI can have high reference values if selected properly.

## Acknowledgement

This work was supported by the Science and Technology Commission of Shanghai Municipality (17JC1401600).